\documentclass[conference]{IEEEtran}
\IEEEoverridecommandlockouts

\usepackage{cite}
\usepackage{amsmath,amssymb,amsfonts}
\usepackage{algorithmic}
\usepackage{graphicx}
\usepackage{textcomp}
\usepackage{xcolor}
\usepackage{url}

\def\BibTeX{{\rm B\kern-.05em{\sc i\kern-.025em b}\kern-.08em
    T\kern-.1667em\lower.7ex\hbox{E}\kern-.125emX}}

\begin{document}

\title{Predictive Failure Detection in Network Hardware Using Thermal Imaging
and Deep Learning with Sensor Fusion}

\author{%
  \IEEEauthorblockN{Ashly Joseph}
  \IEEEauthorblockA{%
    \textit{Independent Researcher}\\
    ashlyjoseph.me@gmail.com}
  \thanks{%
    Author-accepted manuscript. Published as:
    A. Joseph, ``Predictive Failure Detection in Network Hardware Using
    Thermal Imaging and Deep Learning with Sensor Fusion,'' in \emph{2025
    3rd International Conference on Data Science and Network Security
    (ICDSNS)}, Tiptur, India, 2025,
    doi: 10.1109/ICDSNS65743.2025.11168660.
    \par\medskip
    \textcopyright{} 2025 IEEE. Personal use of this material is permitted.
    Permission from IEEE must be obtained for all other uses, in any current
    or future media, including reprinting/republishing this material for
    advertising or promotional purposes, creating new collective works, for
    resale or redistribution to servers or lists, or reuse of any copyrighted
    component of this work in other works.%
  }
}

\maketitle

\begin{abstract}
Unplanned network hardware malfunctions can interrupt services and cause
expensive downtime in data centers. This paper presents a deep learning
approach to predictive maintenance that uses thermal imaging together with
power sensor data to identify early indicators of equipment degradation in
routers, switches, and servers. A simulated dataset was generated comprising
annotated thermal images and power readings representing three operating
states: Normal, Warning, and Critical. Three ImageNet-pretrained convolutional
neural network (CNN) models, ResNet-50, InceptionV3, and VGG16, were assessed
alongside a multi-modal CNN-LSTM fusion model that integrates visual and sensor
time-series information. Experiments were run with and without pre-processing,
namely region-of-interest (ROI) extraction and normalization. Without
pre-processing the CNNs reached only moderate accuracy (ResNet-50 at 52\%),
while ROI-based pre-processing raised ResNet-50 accuracy to 91\%. The CNN-LSTM
model reached the highest accuracy of 94\%, with macro-averaged precision and
recall near 0.95. The results indicate that domain-specific pre-processing and
sensor fusion both contribute substantially to classification performance on
this dataset, providing a starting point for non-intrusive monitoring of
network hardware.
\end{abstract}

\begin{IEEEkeywords}
Predictive Maintenance, Thermal Imaging, Deep Learning, Pre-trained Models,
Network Hardware, Sensor Fusion
\end{IEEEkeywords}

\section{Introduction}
Failures in network hardware in data centers and communication networks can
lead to significant downtime, service interruptions, and financial loss. Many
failures are preceded by physical indicators such as abnormal heat patterns or
irregular power draw. Predictive maintenance, meaning the identification of
early failure indicators before a catastrophic breakdown, has therefore become
important for infrastructure
resilience.~\cite{lee2015, basit2024, it_pms2025}

Thermal imaging provides a non-invasive, real-time technique for monitoring
temperature distributions in hardware, allowing early identification of
problems such as overheated components or a failing power
supply~\cite{ukidwe2023}. Modern data centers also deploy power and temperature
sensors that generate continuous telemetry complementing thermal imagery.
Traditional monitoring systems, however, often rely on static thresholds or
simple rules, which can be inadequate for identifying compound failure signals.

Recent studies have applied machine learning and image processing to failure
detection. Liu et al.~\cite{liu2023} used segmentation and manual feature
extraction on thermal images followed by an SVM classifier to identify server
faults. Such methods rely heavily on domain knowledge and hand-crafted
features. Deep learning, and convolutional neural networks (CNNs) in
particular, offers a data-driven alternative that can learn discriminative
patterns directly from images~\cite{zhao2019}. Transfer learning from
ImageNet-trained models has shown promise in related domains such as electrical
defect detection~\cite{perez2021, ukiwe2024, yosinski2014, krizhevsky2012}.

This paper investigates predictive failure detection in network hardware using
thermal images and power sensor data. Three ImageNet-pretrained CNN models,
ResNet-50, InceptionV3, and VGG16, are assessed, and a hybrid CNN-LSTM model
that integrates the image and time-series modalities is introduced. Model
performance is measured on a simulated dataset labeled into three states
(Normal, Warning, Critical) under two conditions: raw input data, and data
enhanced by region-of-interest (ROI) extraction and normalization.

The contributions of this work are as follows:
\begin{itemize}
  \item A multi-modal dataset of thermal images and power sensor signals for
        network devices is simulated and described.
  \item Three ImageNet-pretrained CNNs and a CNN-LSTM hybrid are compared on
        failure-state classification.
  \item The effect of pre-processing on model performance is quantified,
        reaching up to 39 percentage points of accuracy on this dataset.
  \item The contribution of multi-modal fusion over image-only classification
        is measured.
\end{itemize}

\section{Methodology}

\subsection{Dataset Simulation and Setup}
A multi-modal dataset was generated in simulation to represent a network
equipment rack under normal and degrading thermal conditions. Each simulated
observation consists of a thermal image, which is a two-dimensional temperature
map of a device, paired with a power reading sampled at the same instant.

The simulated configuration comprises three switches, two routers, and four
servers. Thermal image parameters were chosen to match a FLIR A65 class
camera: $640 \times 512$ resolution, 50~mK sensitivity, and a 30~Hz frame rate,
modeled at a standoff of 1.2~m with a $45^{\circ}$ field of view. Power draw
was modeled on outlet-level PDU telemetry sampled at 5-second intervals.

Failure scenarios comprised fan malfunction (impaired airflow), power supply
overload, and CPU stress-related overheating. Sequences were labeled according
to thermal and power trends:
\begin{itemize}
  \item \textbf{Normal (N):} steady-state operation, thermal deviation
        $\leq 5^{\circ}$C from baseline.
  \item \textbf{Warning (W):} gradual hotspot rise of 6--12$^{\circ}$C, or
        power anomalies of $\pm 1$--$2\sigma$ deviation.
  \item \textbf{Critical (C):} rapid thermal rise $> 12^{\circ}$C, or
        shutdown-like power behavior ($> 2\sigma$ deviation or dips).
\end{itemize}

The dataset consists of 900 labeled sequences: 400 Normal, 300 Warning, and 200
Critical. Thermal and power streams share a common timestamp, forming
multi-modal input sequences for training and evaluation.

\subsection{Pre-processing}
Two pre-processing steps were applied before model input: region-of-interest
extraction and feature normalization.

\textbf{ROI extraction.} Each thermal image was cropped to the device's
heat-generating components rather than the full frame, removing rack and room
background. Restricting the input to the device raises the proportion of pixels
carrying state-relevant information and has been reported to improve model
performance in related thermal tasks~\cite{mdpi2024}. ROIs were defined by
fixed bounding boxes per device type, placed around CPU heatsinks and PSU
modules, and held constant across all images of that device type.

\textbf{Normalization.} Thermal pixel intensities, which encode temperature,
were linearly scaled to the range $[0,1]$ and mean-centered per image. Power
readings were z-score normalized by subtracting the mean power draw under
normal operation and dividing by the standard deviation, so that power
anomalies appear as deviations from a common baseline. Normalization matters
particularly for fusion, since it places the two modalities on comparable
scales and prevents either from dominating the learned
representation~\cite{waqas2024, strem2025}.

\subsection{Models}
Three CNN architectures pre-trained on ImageNet were used: ResNet-50,
InceptionV3, and VGG16. ResNet-50 uses residual skip connections to train very
deep networks, InceptionV3 applies parallel convolutional filters of differing
kernel sizes, and VGG16 is a conventional deep CNN of sequential convolutional
layers. The final classification layer of each network was replaced by a fully
connected layer with softmax activation and three outputs (N, W, C), and the
networks were fine-tuned on the dataset for 20 epochs using categorical
cross-entropy loss and the Adam optimizer at a learning rate of $10^{-4}$.

A CNN-LSTM hybrid was built alongside the individual CNNs to use both
modalities. The CNN branch has four 2D convolutional layers with 32, 64, 128,
and 256 filters ($3 \times 3$ kernels, stride 1, padding 1), each followed by
ReLU activation and $2 \times 2$ max-pooling. The resulting feature maps are
flattened and passed through a 128-unit fully connected layer with ReLU and 0.5
dropout. The LSTM branch is a single layer of 50 units with tanh activation
that processes the normalized power series; LSTMs are a standard choice for
anomaly detection over sensor time
series~\cite{malhotra2015, cnn_lstm_rul2024}.

The CNN and LSTM branch outputs are concatenated and passed through a dense
layer for final classification~\cite{atrey2010, cienet2024}. The rationale for
fusion is that different failure modes are more visible in different
modalities~\cite{selfonn2024}: a cooling failure has a pronounced thermal
signature, whereas an electrical overload appears mainly in power draw. The
fusion model used the same loss and optimizer settings as the CNNs, with early
stopping on validation loss.

\subsection{Evaluation Metrics and Procedure}
The dataset was partitioned into training (70\%), validation (15\%), and test
(15\%) subsets. All reported results are on the held-out test set. Standard
multi-class metrics are used: overall accuracy, and precision, recall, and
F1-score derived from the confusion matrix~\cite{sokolova2009}. For a given
class, True Positives (TP) are instances correctly assigned to that class,
False Positives (FP) are instances of other classes assigned to it, and False
Negatives (FN) are instances of that class assigned elsewhere. Recall and
precision are given in Equations~\eqref{eq:recall} and~\eqref{eq:precision}:

\begin{equation}
  \text{Recall} = \frac{TP}{TP + FN}
  \label{eq:recall}
\end{equation}

\begin{equation}
  \text{Precision} = \frac{TP}{TP + FP}
  \label{eq:precision}
\end{equation}

The F1-score is the harmonic mean of precision and recall. In a maintenance
setting, recall on the Critical class matters most, since a missed critical
state is costly, while precision matters for avoiding false alarms. Each model
was evaluated under two conditions: (A) no pre-processing, and (B) ROI
extraction plus normalization. Initialization, training schedule, and
hyperparameters were held constant across conditions.

\begin{figure}[htbp]
\centering
\includegraphics[width=\linewidth]{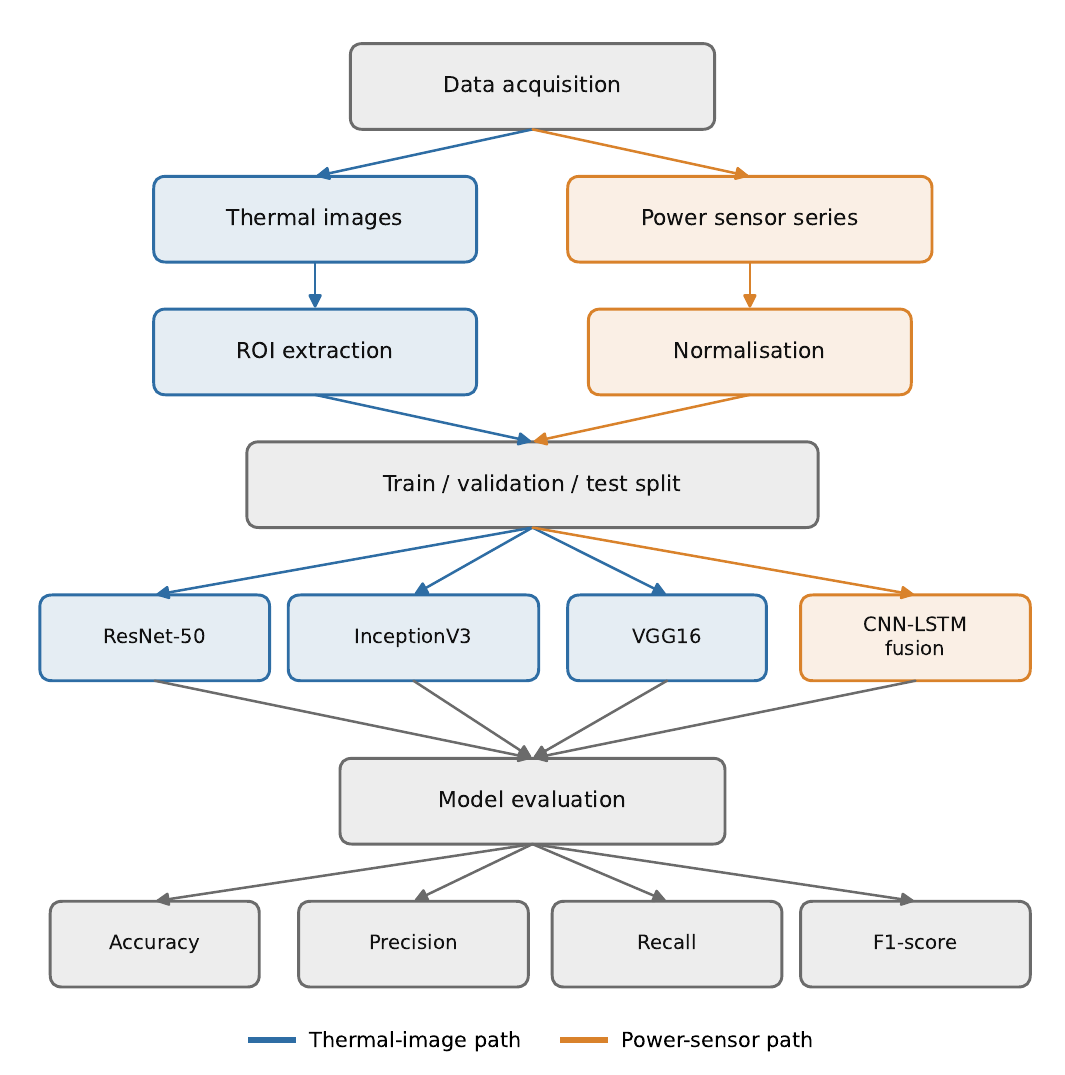}
\caption{End-to-end pipeline. Thermal images and power readings are acquired
in parallel, pre-processed along separate paths, then split and fed to four
models. The three CNNs use the image path only; the CNN-LSTM fusion model uses
both.}
\label{fig:workflow}
\end{figure}

Fig.~\ref{fig:workflow} shows the pipeline. Thermal images and power readings
are acquired in parallel and pre-processed along their own paths, ROI
extraction for images and normalization for sensor values. The processed data
is split into training, validation, and test sets. Four models are trained:
ResNet-50, InceptionV3, VGG16, and the CNN-LSTM fusion model. Evaluation
computes accuracy, precision, recall, and F1-score for each.

\section{Results and Discussion}
Table~\ref{tab:raw} reports results without pre-processing and
Table~\ref{tab:preprocessed} reports results with ROI extraction and
normalization. Both tables give macro-averaged metrics across the three
classes.

\begin{table}[htbp]
\caption{Performance on raw (unprocessed) data}
\label{tab:raw}
\centering
\begin{tabular}{|l|c|c|c|c|}
\hline
Model & Accuracy & Precision & Recall & F1 Score \\
\hline
ResNet-50 & 0.52 & 0.53 & 0.52 & 0.52 \\
InceptionV3 & 0.47 & 0.48 & 0.47 & 0.46 \\
VGG16 & 0.50 & 0.51 & 0.50 & 0.50 \\
CNN-LSTM & 0.60 & 0.61 & 0.59 & 0.60 \\
\hline
\end{tabular}
\end{table}

Without pre-processing the pre-trained CNNs reach 47--52\% accuracy. ResNet-50
and VGG16 sit near 52\%, above the 33\% chance rate for three balanced classes
but well short of usable, and InceptionV3 reaches 47\%. Precision and recall
hover near 0.5, indicating the models separate Warning, Critical, and Normal
poorly from unprocessed images. The CNN-LSTM reaches 60\%, suggesting the
sensor branch contributes even without image pre-processing.

\begin{table}[htbp]
\caption{Performance with ROI extraction and normalization}
\label{tab:preprocessed}
\centering
\begin{tabular}{|l|c|c|c|c|}
\hline
Model & Accuracy & Precision & Recall & F1 Score \\
\hline
ResNet-50 & 0.91 & 0.92 & 0.90 & 0.91 \\
InceptionV3 & 0.88 & 0.88 & 0.87 & 0.88 \\
VGG16 & 0.85 & 0.85 & 0.84 & 0.85 \\
CNN-LSTM & 0.94 & 0.95 & 0.94 & 0.94 \\
\hline
\end{tabular}
\end{table}

All three CNNs improve substantially on ROI-cropped, normalized inputs.
ResNet-50 reaches 91\% accuracy, up from 52\%. InceptionV3 and VGG16 reach 88\%
and 85\%. ResNet-50 is the strongest single-modality model here (F1 = 0.91),
consistent with reports that its deeper residual architecture generalizes well
under transfer learning.

The CNN-LSTM fusion model reaches 94\% accuracy with macro precision and recall
between 0.94 and 0.95. The image-only CNNs must infer state from appearance
alone, while the fusion model can also use the power trend, which plausibly
accounts for the gap in ambiguous cases. This is an inference from the
aggregate numbers rather than a measured attribution; an ablation that feeds
the power series to a classifier on its own would separate the two
contributions.

Comparing Tables~\ref{tab:raw} and~\ref{tab:preprocessed}, accuracy rises by
roughly 35--39 percentage points across models. This is consistent with
findings in thermal diagnostics in other domains, where restricting the input
to the relevant region and normalizing intensity are reported as material
preprocessing steps. One plausible mechanism is that in an uncropped frame most
pixels carry no state information, so the convolutional filters allocate
capacity to background variation; this remains a hypothesis and is not tested
here.

\section{Conclusion}
This paper compares pre-trained deep learning models for early identification
of failure states in network hardware using thermal images and power sensor
data, with and without ROI extraction and normalization. On this simulated
dataset, pre-processing raised the best CNN (ResNet-50) from roughly 52\% to
91\% accuracy, and the CNN-LSTM fusion model reached 94\% by combining thermal
and power inputs. The results indicate that on this data, input preparation
affects classification accuracy at least as much as architecture choice.

Practically, operators building thermal predictive maintenance pipelines should
budget effort for pre-processing: automatic device detection and cropping,
temperature range calibration, and sensor noise handling, before model
selection. Combining thermal imagery with power or fan telemetry also improved
results here~\cite{atrey2010, cienet2024}.

Several extensions follow. First, the method should be evaluated on empirical
data from operating network equipment to test whether the simulated results
transfer, using infrared cameras on live routers and switches under induced
failure conditions. Second, stronger architectures could be explored, including
object detection models that locate hotspots automatically and transformer
models over multi-modal
sequences~\cite{selfonn2024, wavelet2024, gertsvolf2024}. Third, expanding the
label set beyond Normal, Warning, and Critical to name specific fault types
(fan failure, PSU fault) would be more actionable and may require multi-label
or hierarchical models. Fourth, deploying on edge hardware for real-time
inference, using quantization or knowledge distillation to compress the
CNN-LSTM model, is a practical next step.

\end{document}